\documentclass[conference]{ieeeconf}
\IEEEoverridecommandlockouts

\usepackage{amsmath,bm,amssymb,amsfonts}
\usepackage{algorithmic}
\usepackage{graphicx}
\usepackage{textcomp}
\usepackage{xcolor}
\usepackage{epsfig}
\usepackage{gensymb}
\usepackage{subfig}
\usepackage{float}
\usepackage{booktabs}
\usepackage{multirow}
\usepackage{makecell}
\usepackage{hyperref}
\usepackage[linesnumbered,boxed,ruled,commentsnumbered]{algorithm2e}
\DeclareMathOperator*{\argmax}{arg\,max}
\DeclareMathOperator*{\argmin}{arg\,min}
\usepackage{cite}
\def\BibTeX{{\rm B\kern-.05em{\sc i\kern-.025em b}\kern-.08em
    T\kern-.1667em\lower.7ex\hbox{E}\kern-.125emX}}
\begin{document}

\title{\LARGE \bf ACSC: Automatic Calibration for Non-repetitive Scanning Solid-State LiDAR and Camera Systems\\
}
\author{Jiahe Cui$^{1}$$^{2}$$^{3}$, Jianwei Niu$^{1}$$^{2}$$^{3}$, Zhenchao Ouyang*$^{3}$, Yunxiang He$^{3}$ and Dian Liu$^{3}$  
\thanks{*Email: ouyangkid@buaa.edu.cn}
\thanks{$^{1}$State Key Laboratory of Virtual Reality Technology and Systems, Beihang University, Beijing 100191, China}%
\thanks{$^{2}$Beijing Advanced Innovation Center for Big Data and Brain Computing (BDBC), Beihang University, Beijing 100191, China}%
\thanks{$^{3}$Hangzhou Innovation Institution, Beihang University, Hangzhou 310000, Zhejiang, China}%
}


\maketitle
\thispagestyle{empty}
\pagestyle{empty}

\begin{abstract}
	Recently, the rapid development of Solid-State LiDAR (SSL) enables low-cost and efficient obtainment of 3D point clouds from the environment, which has inspired a large quantity of studies and applications. However, the non-uniformity of its scanning pattern, and the inconsistency of the ranging error distribution bring challenges to its calibration task. In this paper, we proposed a fully automatic calibration method for the non-repetitive scanning SSL and camera systems. First, a temporal-spatial-based geometric feature refinement method is presented, to extract effective features from SSL point clouds; then, the 3D corners of the calibration target (a printed checkerboard) are estimated with the reflectance distribution of points. Based on the above, a target-based extrinsic calibration method is finally proposed. We evaluate the proposed method on different types of LiDAR and camera sensor combinations in real conditions, and achieve accuracy and robustness calibration results. The code is available at \textcolor[RGB]{236,0,140}{\url{https://github.com/HViktorTsoi/ACSC.git}}. 
\end{abstract}


\section{Introduction}
Multi-sensor fusion has always been essential to robotics and self-driving systems, for accurate perception of the surrounding environment. Among different types of fusion, the commonest one is the combination of LiDAR (Light Detection And Ranging) and the optical camera, especially in the tasks of vision-based odometry and mapping\cite{debeunne2020review}, object detection and tracking tasks\cite{dong2020fast,qi2020imvotenet, ouyang2018multiview, jaritz2020xmuda}. In fusing the multi-modal data from LiDAR-camera systems, the first and most critical step is the accurate extrinsic calibration. The general calibration process is to detect the multiple corresponding 3D-2D corners, and then solve the relative pose between LiDAR and the camera by utilizing the PnP (Perspective-N-Point) method. Thus, it is essential to find the corresponding features accurately by geometric and texture constraints from point clouds and images. 


\begin{figure}[!htb]
	\centering
	\includegraphics[trim=20 60 30 20, clip, width=0.47\textwidth]{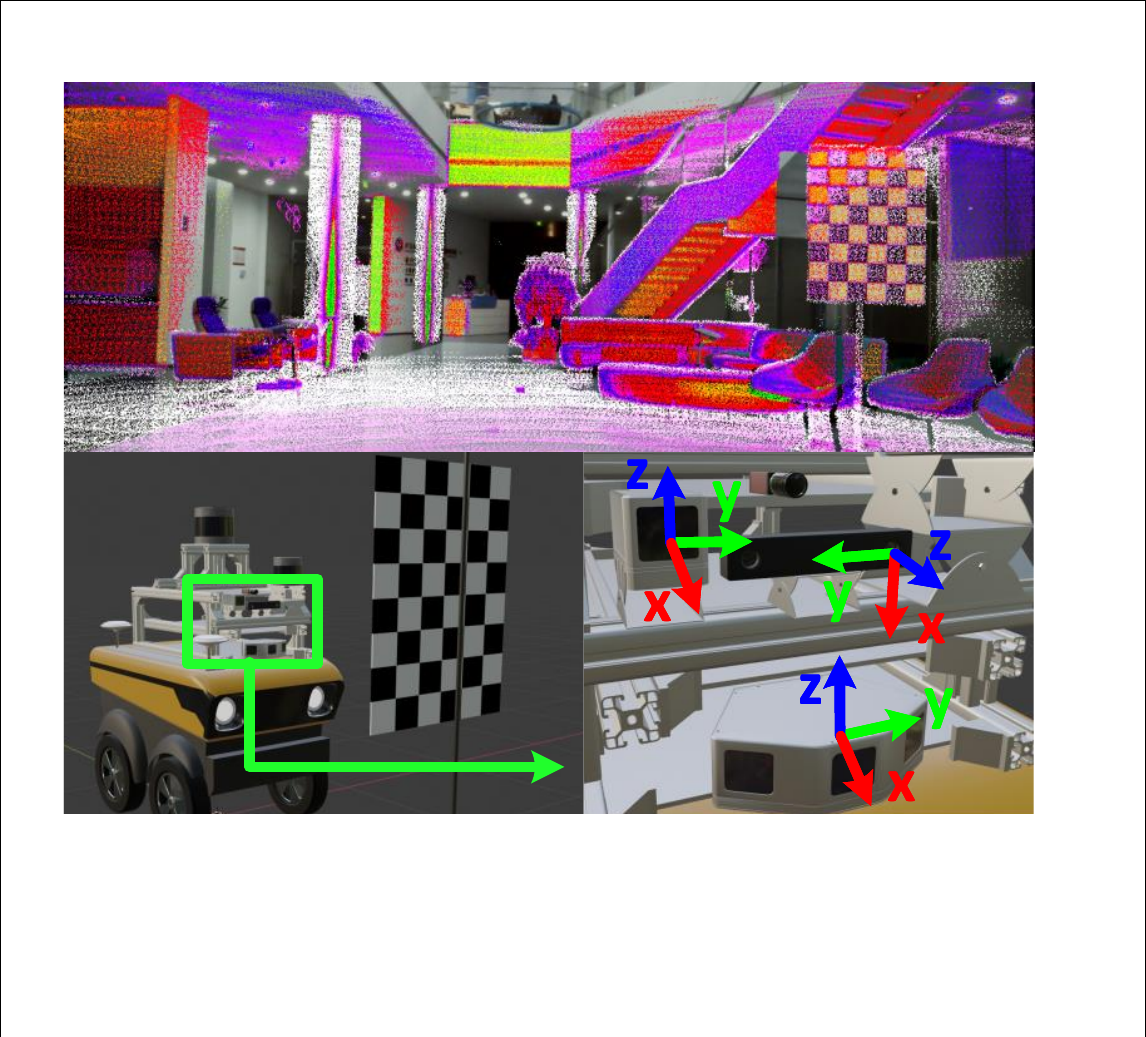}
	\caption{(a) Top: The reprojection of integrated point clouds to the image using the extrinsic parameter solved by ACSC; (b) Bottom: The calibration target we use is a printed checkerboard; the placement is shown in the left; the sensor setup and its coordinate system are described in the right.}
	\label{fig:coord}
	\vspace{-0.4cm}
\end{figure}

In the past two years, the Solid-State LiDAR (SSL) system has begun to be widely adopted by different intelligent unmanned platforms\cite{lin2020loam,liu2020balm, lin2019fast}, due to its low price, automotive-grade design, and similar ranging performance as high-definition LiDARs. However, The characteristic of SSL brings new challenges to sensor deployment, especially for the sensor calibration process: because the non-repetitive scanning mode is utilized by this LiDAR model, the spatial distribution of points is nonuniform; its ranging measurement is sensitive to the texture and color of targets; and its ranging error is also non-uniformly distributed due to its low-cost measuring unit. Such drawbacks seriously affected corresponding feature extraction in point clouds, and thus reduce the performance of automatic calibration. There are few former studies on automatic calibration of SSL-camera systems. 

To solve the above-mentioned problems, this paper proposes ACSC, an Automatic extrinsic Calibration method for the SSL-Camera systems. We first design a time-domain integration and point cloud feature refinement pipeline to extract as much effective information as possible for non-repetitive scanning point clouds, and proposed a 3D corner extraction method by utilizing the reflectance intensity distribution of the calibration target. Based on the 3D corners and corresponding 2D corners(from the optical image), a target-based calibration method is proposed. We also evaluated the proposed method in different types of SSL, e.g., the Livox Mid-100/40 and Horizon SSL, and camera combinations. The main contributions of this paper are summarized as follows:

\begin{itemize}{}
	\item An automatic target-based calibration method for the SSL and camera system is proposed;
	\item A temporal-spatial-based geometric feature refinement, and reflectance intensity distribution-based 3D corner estimation pipeline for point clouds from non-repetitive scanning SSL is presented;
	\item To evaluate the calibration performance, multiple real-world experiments based on various combinations of LiDARs with cameras are deployed, and the proposed method can successfully solve extrinsic parameters from various types of LiDAR-camera systems.
\end{itemize}



\section{Related Works}
Most of the extrinsic calibration methods are based on reference targets, the common idea is to design calibration targets or corresponding features that can be clearly detected in all sensor FOVs. The planar board is the most commonly used calibration target for both intrinsic and extrinsic calibrations of the monocular/stereo cameras\cite{fuersattel2017accurate, gu2019environment, prokos2012automatic, huang2020improvements}. Fremont et al.\cite{fremont2008extrinsic} designed a black circle-based planar board, then searched the 3D coordinates of the circle center and the normal vector of the plane for extrinsic calibration between a camera and a LiDAR. Zhou et al.\cite{zhou2018automatic} proposed a single-shot calibration method, by extracting line features from the LiDAR points and the image. Wang et al.\cite{wang2017reflectance} proposed a 3D corner estimation algorithm based on the correlation between the reflectance intensity of the laser and the color of the calibration target.

The targetless-based method tries to find natural patterns (mainly the line or orthogonal) from the scene, and formulates them in terms of geometric constraints to then solve the extrinsic. Basically, they can be divided into static-based and motion-based methods. The former one is similar to target-based registration, finding reference features with static targets. Scaramuzza et al.\cite{scaramuzza2007extrinsic} proposed an extrinsic calibration technique that requires manually associating 2D points on the image with 3D points on the point cloud. In \cite{gong20133d}, the trihedral features detected from the environment, such as walls and corners on the street, are used for calibration, however, the method relies on high-cost Velodyne-64E to find the trihedron patterns. \cite{pandey2015automatic} proposed a targetless calibration method, the extrinsic parameter is solved by maximizing the mutual information (MI) obtained between the sensor-measured surface intensities from images or point clouds. In contrast, motion-based methods\cite{fu2019lidar} use Simultaneous Localization And Mapping (SLAM) technologies, the extrinsic is calculated based on the motion estimation between the fixed sensors, by minimizing the trajectory to trajectory matching error.

The methods above have achieved fine results on mechanical LiDAR and camera systems, but it is challenging to generalize them to SSL, whose point cloud measurement have significantly different feature distribution from the mechanical models. Scaramuzza et al.\cite{garcia2020geometric} proposed a geometric model-based method for SSL intrinsic correction. Liu et al.\cite{lin2020decentralized} proposed a multi-SSL calibration method by utilizing LOAM-based trajectory matching. The intensity value is related to the superficial texture and color of the measured targets; therefore, it can provide as much plentiful semantic information as optical images, especially for dense point cloud from non-repetitive scanning SSL after time-domain stacking.  In this paper, by utilizing temporal integration and feature refinement, the proposed calibration method can make the best of the non-repetitive pattern of SSL, and achieve accuracy calibration results. 

\vspace{-0.5cm}
\begin{figure}[H]
	\centering
	\subfloat[single scan]{\includegraphics[trim=0 15 0 75, clip, width=0.16\textwidth]{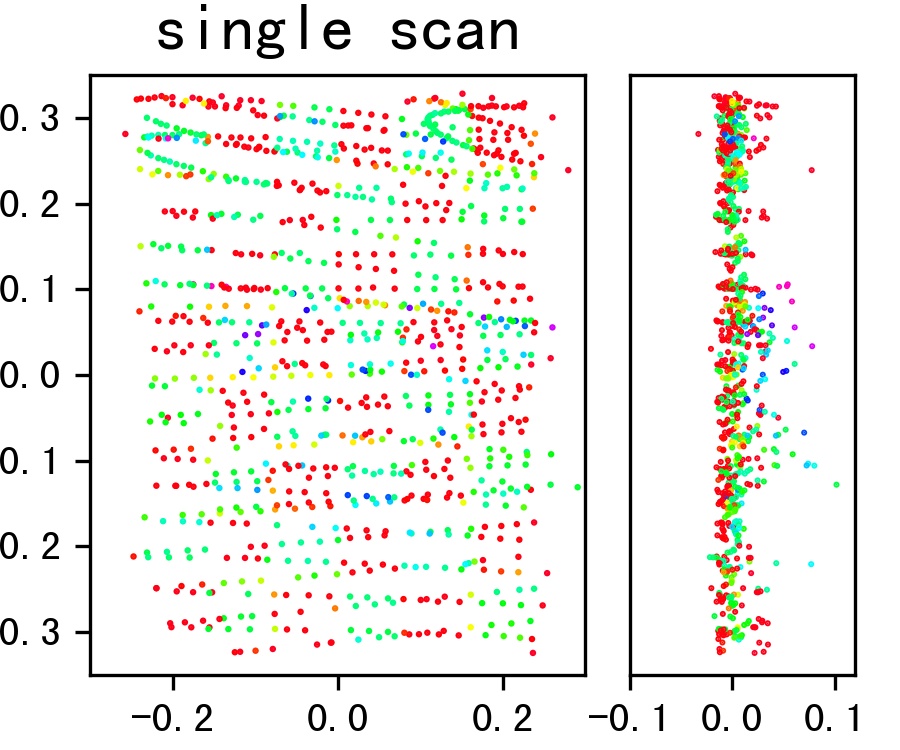}}
	\subfloat[distance: 1.5m]{\includegraphics[trim=0 15 0 75, clip, width=0.16\textwidth]{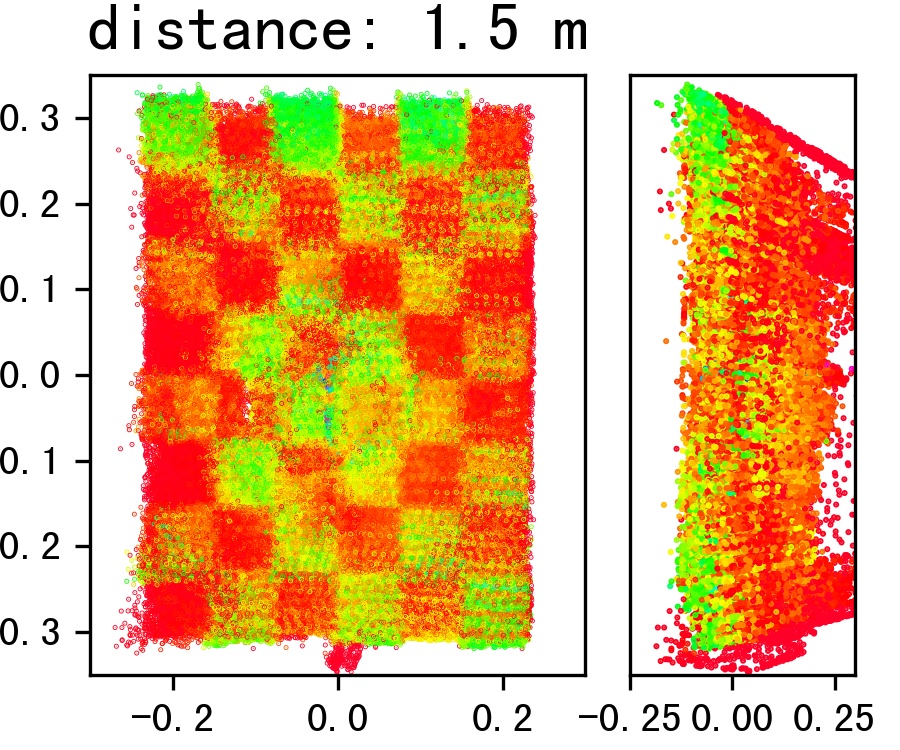}}
	\subfloat[distance: 3.9m]{\includegraphics[trim=0 15 0 75, clip, width=0.16\textwidth]{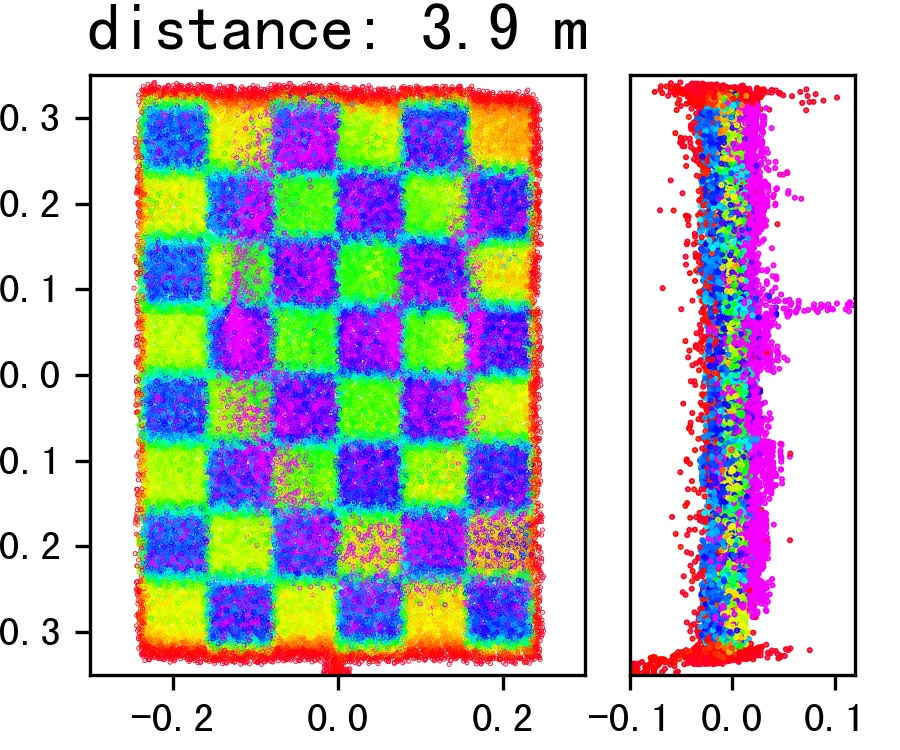}}
	\\[-2ex]
	\subfloat[distance: 7.6m]{\includegraphics[trim=0 15 0 75, clip, width=0.16\textwidth]{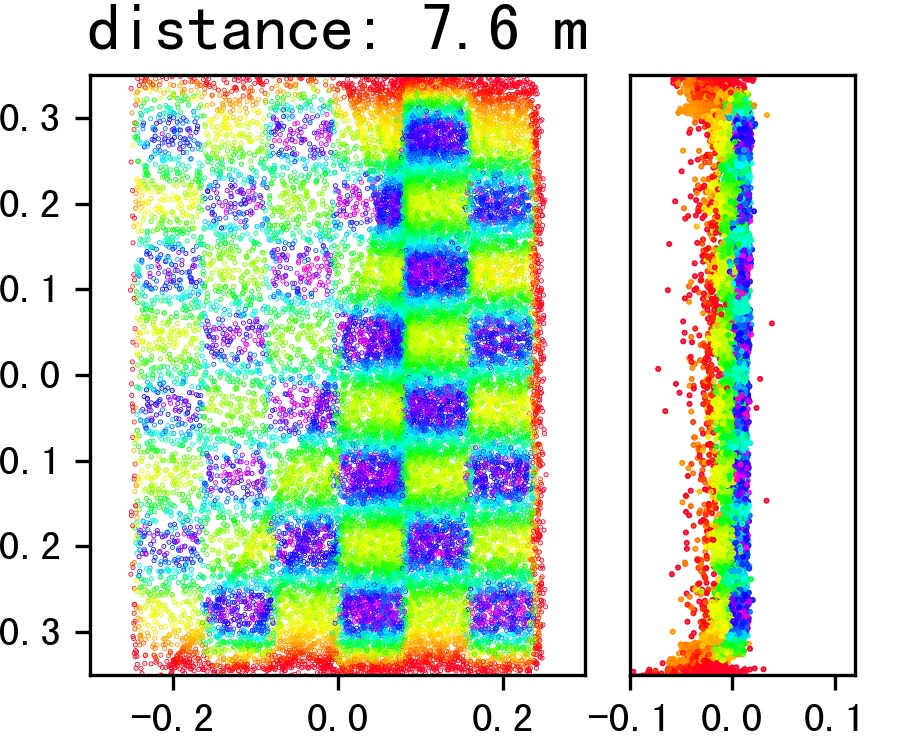}}
	\subfloat[distance: 8.7m]{\includegraphics[trim=0 15 0 75, clip, width=0.16\textwidth]{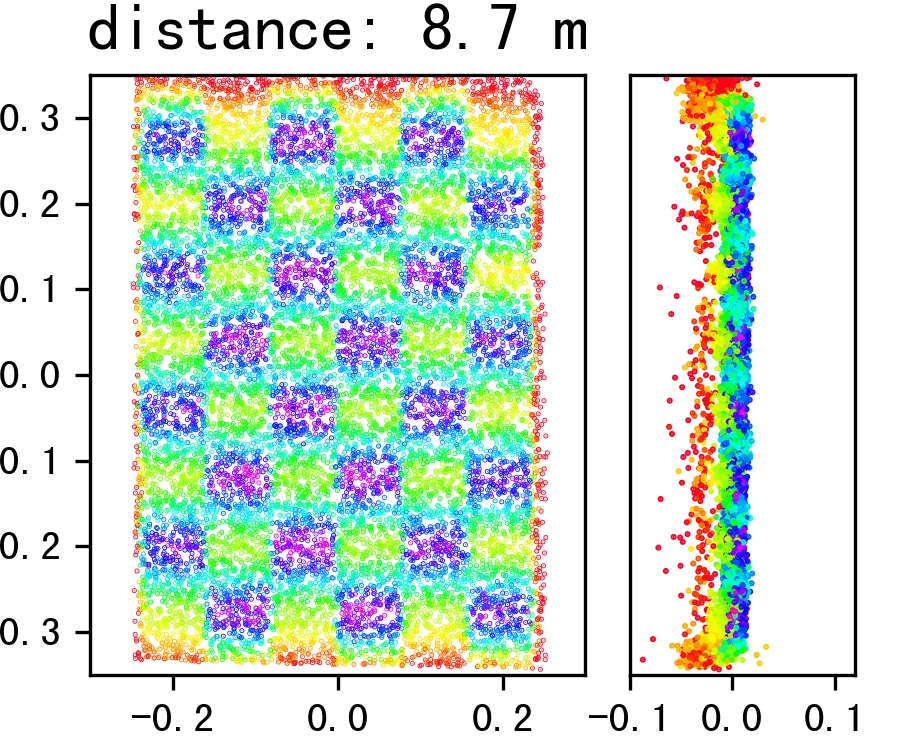}}
	\subfloat[distance: 10.7m]{\includegraphics[trim=0 15 0 75, clip, width=0.16\textwidth]{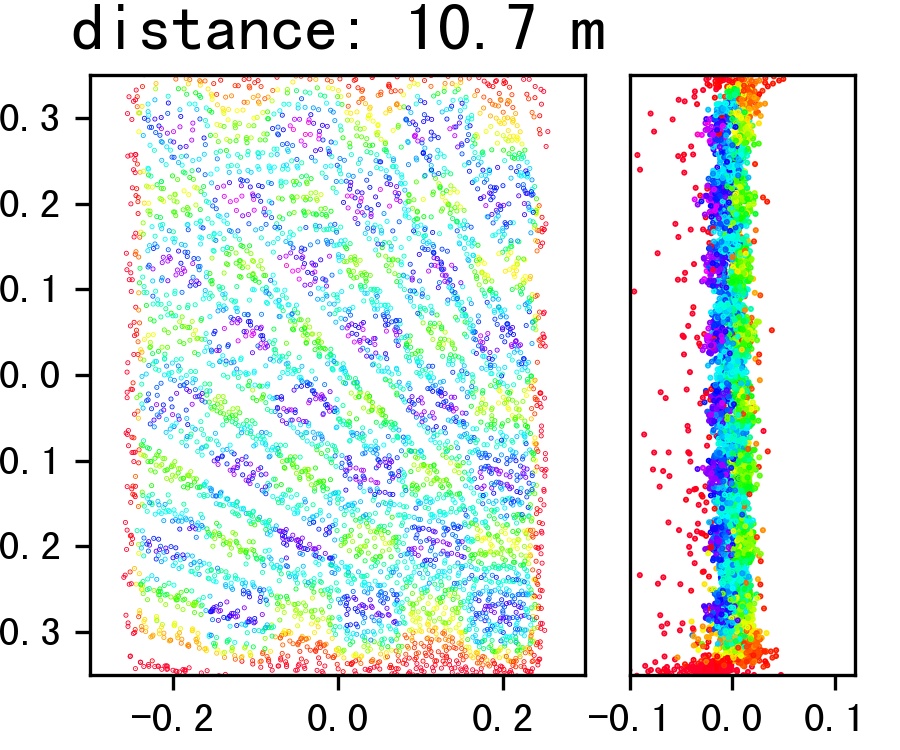}}
	\\[-1ex]
	\caption{The integrated (without feature refinement) checkerboard point clouds from non-repeat scanning SSL (left: front view; right: side view).}
	\label{fig:scanning_pattern} 
	\vspace{-0.3cm}
\end{figure}

\begin{figure*}[!htb]
	\setlength{\abovecaptionskip}{-0.cm}
	\setlength{\belowcaptionskip}{-0.cm}
	\centering
	\includegraphics[trim=20 270 75 12, clip, width=0.99\textwidth]{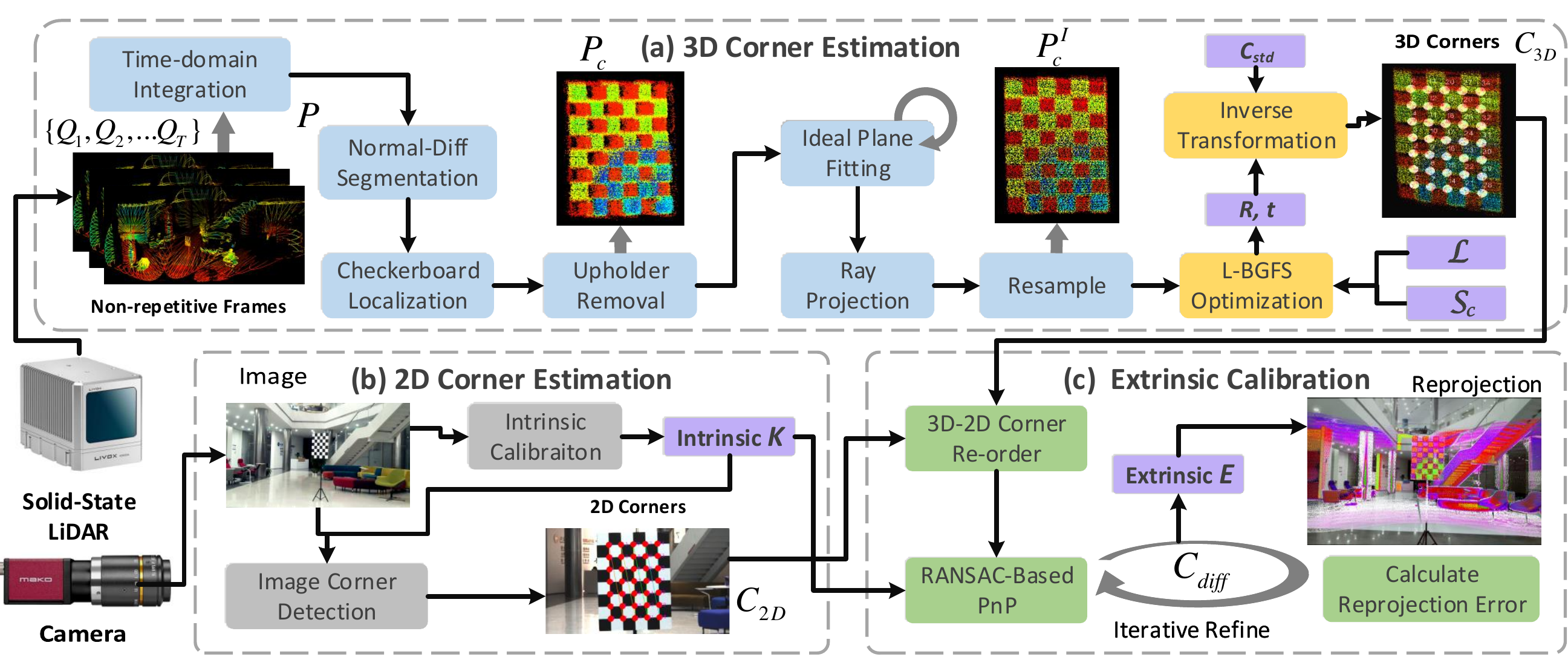}\\
	\caption{The proposed calibration method. (a) The feature refinement and checkerboard inner corner estimation process of point clouds from SSL. The input $\{Q_1,Q_2,...,Q_T\}$ are incoming frames of SSL, $S_c$ represents the standard model constructed from the geometric parameter of the calibration target we use, $C_{std}$ is the corner generated from $S_c$, and $\mathcal{L}$ is the similarity measurement function for optimizing the 3D corner location; (b) 2D corner estimation from images; (c) The extrinsic calibration process. }
	\label{fig:pipeline}
	\vspace{-0.4cm}
\end{figure*}
\section{Method}
\label{sec:methods}
For the SSL-camera system, the problem of extrinsic calibration is to estimate the relative rotation and translation between the two sensors, namely, the task is to solve the extrinsic parameter matrix ($\bm E\in SE3$) based on the corresponding 3D-2D corners extracted from same frame of the two different sensors, respectively. The proposed method uses a printed checkerboard as the calibration target, the dimension of the inner corners on the board is $N_h \times N_w$, and the inner grid size is $G_s$. 

However, the main challenge is how to accurately extract the corners from the unstably-distributed point clouds. We conduct research on the representative Livox series LiDAR in this paper, and Fig.~\ref{fig:scanning_pattern} shows several typical patterns obtained in scanning of the calibration target: 1) The non-repetitive scanning pattern causes sparse single frame measurement. Although the point cloud can be densified by multiple-frame stacking, the outliers are also reserved (Fig.~\ref{fig:scanning_pattern} (a)-(f) ). 2) The range measurement has large variance in axial direction (direction of beams). The closer the distance, the more serious the jitter (side views in  Fig.~\ref{fig:scanning_pattern}). 3) The special scanning pattern also leads to uneven distribution of scan lines (Fig.~\ref{fig:scanning_pattern} (d), (e) and (f)). The  proposed calibration method considers the above characteristics, and thus can accurately extract corresponding feature from SSL and camera. The whole workflow is illustrated in Fig.~\ref{fig:pipeline}.


\subsection{Calibration Target Feature Refinement}
\label{sec:feature_refinement}

\subsubsection{Time-domain integration of point clouds}
We take advantage of the non-repetitive scanning pattern, and integrate the continuous scans in the time domain to densify the point clouds, instead of using single scan Fig.~\ref{fig:scanning_pattern} (a) directly. In this way, the dense checkerboard measurement can be obtained, with intensity features that provides as much semantic information as the image does. However, if incoming point clouds are simply stacked, the noise points from each frame will also be accumulated, and eventually cause fuzzy results (as shown in Fig.~\ref{fig:scanning_pattern}); therefore, we first utilized statistical outliers removal for every incoming frame, based on the neighbor density distribution of points, the integrate the noise-free points in time-domain, the detailed process is shown in Algorithm. \ref{alg:integration}. 

After integration, the surface normal of the points is distributed more continuously than that of a single scan, therefore, the normal difference-based segmentation\cite{ioannou2012difference} is first deployed to extract the candidate clusters that may contains the checkerboard. We then sort the clusters by the standard target similarity measurement $\mathcal{L}$ (presented in \textbf{Section}~\ref{sec:3d_corner_estimation}), to measure the difference between the cluster and the calibration target, and only the clusters with the minimum difference are maintained, as the located checkerboard measurement (marked as $\bm{P_c}$).



\IncMargin{1em} 
\begin{algorithm}[!ht]
	\SetAlgoNoLine
	\SetKwInOut{Input}{\textbf{Input}}\SetKwInOut{Output}{\textbf{Output}}
	\Input{\\
		$T$ frames of point cloud $\bm Q=\{Q_1,Q_2,...,Q_T\}$\; \\
		Threshold of neighbor size $K$;\\
		Scaling factor $scale_{std}$;}
	\Output{\\
		Stacked point cloud $\bm P$\; }
	\BlankLine
	Initialize $P = \emptyset $\;
	\Repeat
	{\text{All $T$ frames of point clouds in $\bm Q$ are consumed}}
	{
		for frame $Q_t$, build KD-Tree\;
		\For {each point $q_i \in Q_t$}{
			search the $K$ nearest points of $q_i$\;
			calculate the average distance ($disK_i$) of the $K$ neighbours\; 
		}
		calculate the mean distance $meanK$ and standard deviation $stdK$ of $[disK_0, disK_1, ..., disK_N]$\;
		$thresh = meanK + scale_{std} \times stdK $\;
		\For {each point $q_i \in Q_t$}{
			\uIf{$ disK_i < thresh $} {
				Remove $q_i$ from $Q_t$\;
			}
		}
		$\bm P = \bm P + \{ Q_t \}$\;
	}
	\caption{The time-domain integration process of incoming point cloud frames from SSL.}
	\label{alg:integration}
\end{algorithm}
\DecMargin{1em}

\newlength{\textfloatsepsave} 
\setlength{\textfloatsepsave}{\textfloatsep} 
\setlength{\textfloatsep}{10pt}

\subsubsection{Feature refinement}
\begin{figure}[!htb]
	\centering
	\includegraphics[trim=20 505 410 25, clip, width=0.48\textwidth]{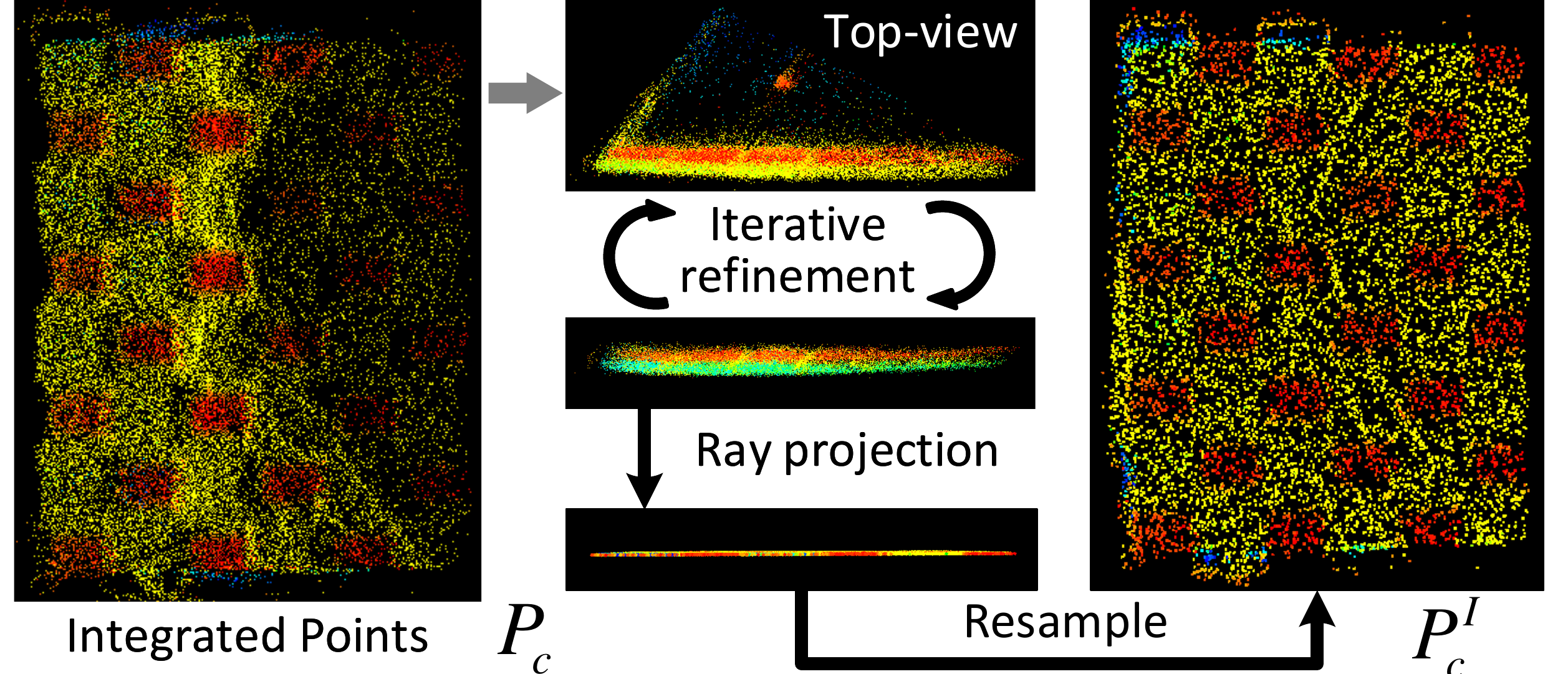} 
	\caption{The feature refinement process, from raw point cloud of the checkerboard $P_c$, which is segmented from background points, to the noise-free measurement $P^I_c$.}
	\label{fig:refine} 
	\vspace{-0.3cm}
\end{figure}

To obtain a noised-free and accurate checkerboard measurement, a series of point cloud feature enhancement methods are designed, and the whole process is shown in Fig.~\ref{fig:refine}.

Considering the upholder/ground points may be connected to the checkerboard $P_c$ after being segmented from the background, we first remove these points. Suppose the placement of the sensor and the checkerboard are as shown in Fig.~\ref{fig:coord}, only the upper and lower boundaries of the checkerboard points in the Z-axis direction are needed to be determined. For the upper boundary, we choose the $Z$ coordinate of the highest point $Z_{top} = \max{P^z_c}$. The lower boundary is calculated based on both the width distribution $H$ of checkerboard points along Z-axis, and the checkerboard size constraint:

\vspace{-0.5cm}
\begin{equation}
	\label{eq:lowboundary}
	\centering
	\begin{aligned}
		Z_{down}=\argmax_{z}                                           
		{                                                              
		\frac                                                          
		{\partial {H(|| P^{xy}_c - \frac{1}{N} \sum P^{xy}_c ||, z)} } 
		{\partial {z}}                                                 
		}                                                              
		                                                               \\
		+ ~                                                            
		w \frac                                                        
		{| (Z_{top} - z) - N_h G_s |}                                  
		{N_h G_s}                                                      
	\end{aligned}
\end{equation}
Where $w$ is the ratio of checkerboard size constraint. All the other points that are not in this range $(Z_{down}, Z_{top})$ will be removed. For simplicity, the remained checkerboard point cloud is also marked as $P_c$.

As shown in Fig.~\ref{fig:scanning_pattern}, the point clouds are scattered along the axial direction due to due systematic error of SSL measurement, while in the ideal situation, all points are supposed to fall on the checkerboard plane, therefore, it is hard to estimate the accurate corner coordinates. We thus designed an iteratively refining method to fit the point cloud to the ideal plane where the checkerboard lies. During each iteration, we first use the RANSAC-based method to fit the optimal plane. The point clouds ${P'_c}$ within a distance of $\sigma$ to the plane are reserved: 

\vspace{-0.5cm}
\begin{equation}
	\label{eq:fitting}
	\centering
	\bm {P'_c} = 
	\{
	(x_0,y_0,z_0) \in \bm {P_c}
	\mid
	\frac{|Ax_0+By_0+Cz_0+D|}{\sqrt{A^2 + B^2 + C^2}} < \sigma
	\}
\end{equation}

Here, the $A$, $B$, $C$ and $D$ are the estimated parameter of the fitted plane. Now the point set ${P'_c}$ is closer to the ideal plane than $P_c$, we then shrink $\sigma$ and repeat the fitting and filtering process. When all the point clouds are within the threshold of the current iteration, the calculation stops. According to the ranging principle of SSL, the ranging measurement of a laser beam may contain error in the radial direction (from the obstacle to the sensor center), while that at the azimuth or polar directions can be ignored. Based on this prior, the projective transformation model are then utilized for projecting the noise point cloud onto the ideal plane, as described in Eq.~\ref{eq:unify}.

\vspace{-0.4cm}
\begin{equation}
	\label{eq:unify}
	\centering
	\bm{P^I_c} = \{ (x_0t, y_0t, z_0t) \mid (x_0,y_0,z_0) \in \bm {P'_c} \} 
\end{equation}
where
\begin{equation}
	\label{eq:unify_sub}
	\centering
	t = -D / (Ax_0 + B y_0 + C z_0) 
\end{equation}

Afterwards, to reduce the influence of non-uniformity caused by the non-repetitive scanning pattern (as shown in Fig.~\ref{fig:scanning_pattern}~(d) and (f)), we divide the unified points $\bm {P^I_c}$ is into multiple grids, and then adopt random sampling to each gird, to maintain the point cloud density not greater than $\delta_{\rho}$. By combining the downsampled grid together, we can get a checkerboard point cloud with uniform density. This operation can reduce the impact of high-density areas on penalty function during later corner fitting, and lead to better overall corner detection results.



\setlength{\textfloatsep}{\textfloatsepsave}

\subsection{3D Corner Estimation}
\label{sec:3d_corner_estimation}

\begin{figure}[!htb]
	\centering
	\includegraphics[trim=50 360 340 30, clip, width=0.48\textwidth]{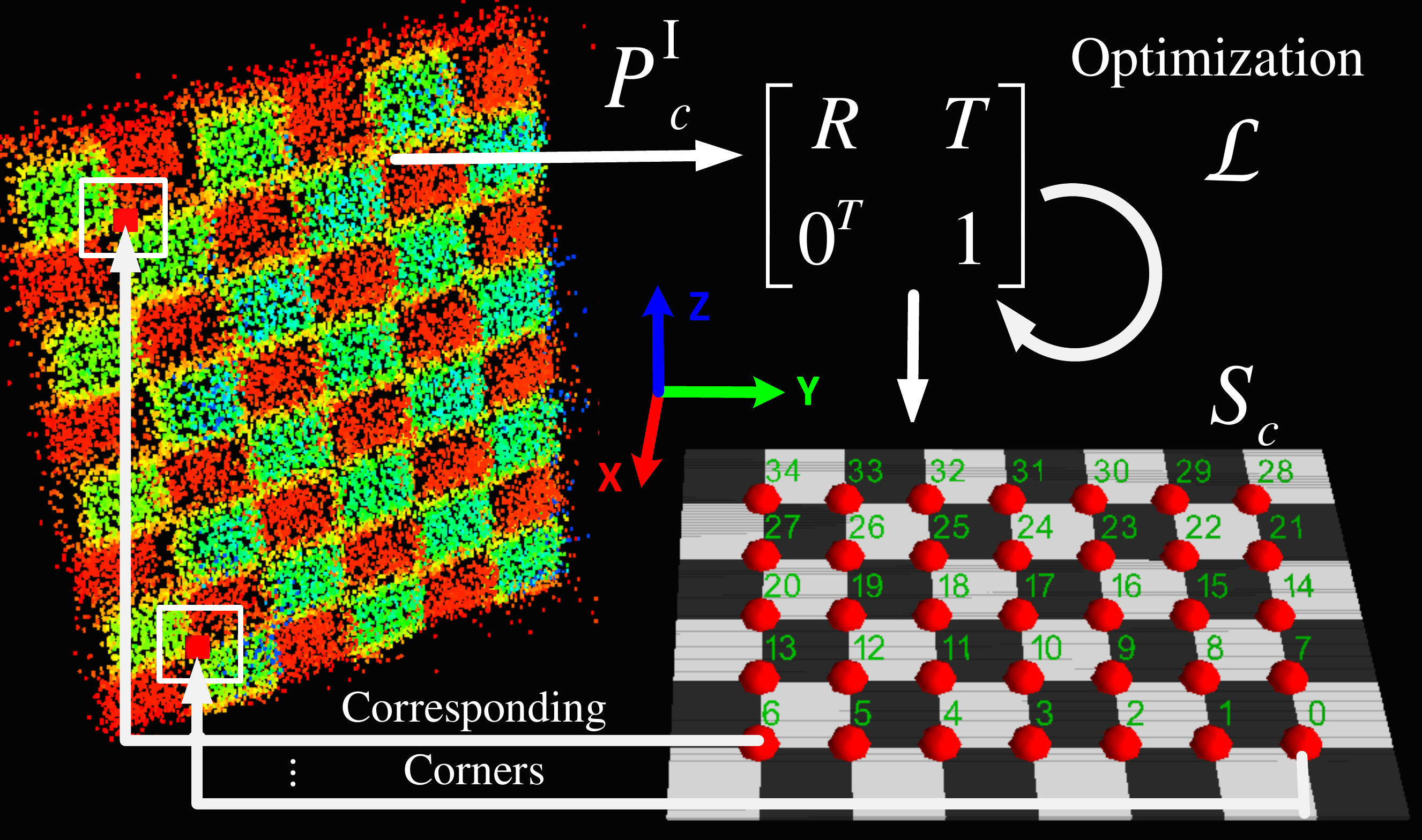} 
	\caption{Schematic diagram of corner estimation. The transformation from the refined checkerboard measurement $P^I_c$ to the standard model $S_c$ is first solved using the reflectance distribution, and then used to inversely transform the corners of the $S_c$ to original checkerboard.} 
	\label{fig:corner} 
	\vspace{-0.4cm}
\end{figure}


We introduce a non-linear optimization method, that is constrained by the global reflectance distribution of checkerboard measurement, for estimating the corners, the schematic diagram is shown in Fig. \ref{fig:corner}. Given the geometrical parameter of the checkerboard $\{N_w, N_h, G_s\}$, a standard checkerboard model $\bm {S_c}$ with reflectance information can be constructed as Eq.\ref{eq:standard_chessboard}:

\vspace{-0.4cm}
\begin{equation}
	\label{eq:standard_chessboard}
	\centering
	\begin{aligned}
		\bm {S_c} = \{ (x,y,z,I) \mid x \in (0, N_w G_s),y \in (0, N_h G_s), 
		\\
		z=0, I=\delta(x,y,z) \}
	\end{aligned}
\end{equation}

Here, $\delta$ is a binary function that indicates the pattern of checkerboard at $(x,y,z)$. Considering the measured value of the checkerboard point cloud ($\bm P^I_c$) by LiDAR is physically the same size with $\bm S_c$ without scaling, the checkerboard measurement and the standard model can be aligned through rigid body transformation, and the 3D inner corner coordinates can also be solved by inverse transformation, since the corners of the standard model are already known from geometrical parameters. Therefore, we convert the 3D corner estimation into a non-linear optimization problem: solving the transformation parameters $\{R,t\}$ that make the measured value of the transformed checkerboard point cloud closest to the ideal standard model.

\begin{figure}
	\centering
	\includegraphics[trim=10 0 0 -11, clip,width=0.49\textwidth]{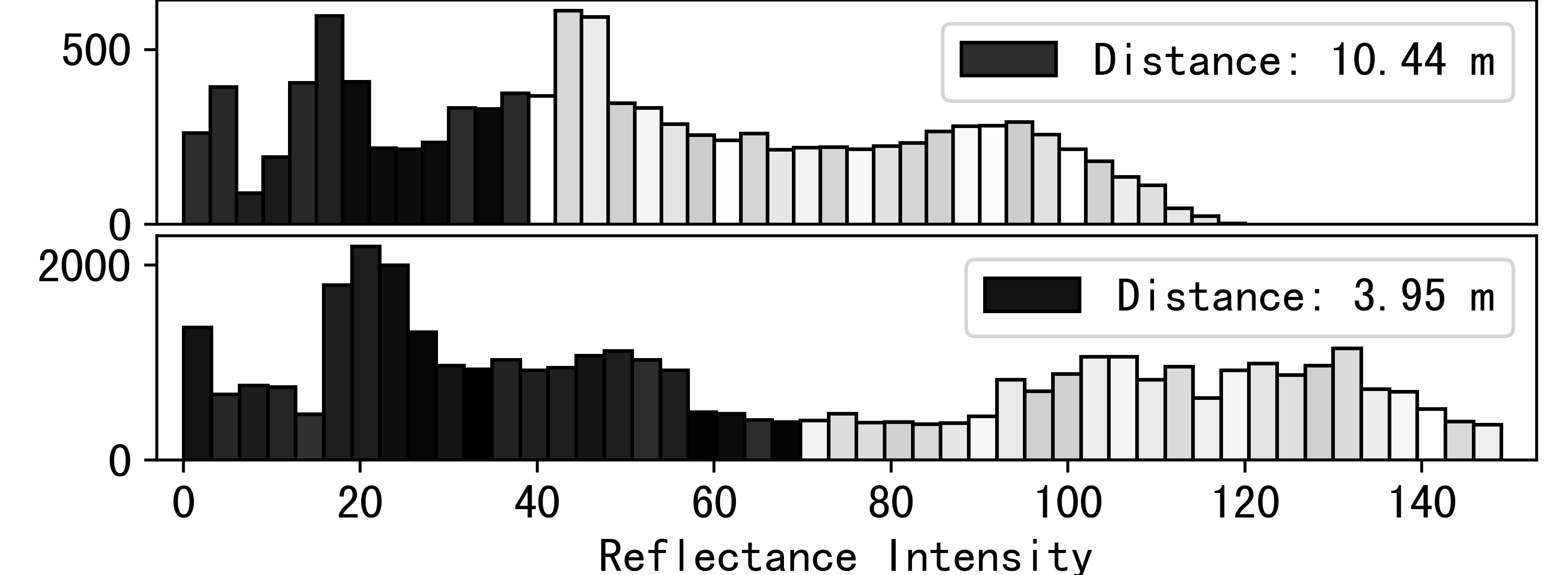} 
	\caption{Reflectivity distribution of checkerboard point cloud under different distances. The black and white bars respectively represent the reflectivity of the corresponding color grid in the checkerboard.}
	\label{fig:intensitydis} 
	\vspace{-0.5cm}
\end{figure}

The critical step is designing the similarity function, that can accurately evaluate the pose-difference between the measured value $\bm P^I_c$ and the standard model $\bm S_c$. We found that, as shown in Fig.~\ref{fig:intensitydis}, the reflectance of the checkerboard measurement $P_c$ and the black-white pattern on the physical checkerboard basically shows the same spatial distribution, therefore, we formulates the similarity as the reflectance distribution difference, defined as Eq.~\ref{eq:similarity_function}:

\vspace{-0.5cm}
\begin{equation}
	\label{eq:similarity_function}                  
	\begin{aligned}
		\mathcal{L}                                         
		(                                                   
		\bm S^{xyz}_c,                                      
		\tilde {\bm P}^{xyz}_c,                             
		\bm I_S,                                            
		\bm I_P
		|
		R, t                                             
		)                                                   
		=                                                   
		\sum                                                
		_{\tilde{p_i} \in \bm{\tilde{P_c}} }                
		^{N_c}                                              
		{\mathcal{L}_1 + \mathcal{L}_2},                    
		                                                    \\
		\mathcal{L}_1                                       
		=                                                   
		\delta_{in}( \tilde{p_i}, \bm S^{xyz}_c )           
		| \bm I_S( \tilde{p_i} ) - \bm I_P( \tilde{p_i} ) | 
		d( \tilde{p_i}, G_i ),                              
		                                                    \\
		\mathcal{L}_2                                       
		=                                                   
		( 1 - \delta_{in}( \tilde{p_i}, \bm S^{xyz}_c ) )   
		d( \tilde{p_i}, G_i )                               
	\end{aligned}
\end{equation}

Here, $\tilde {\bm P}^{xyz}_c$ is the transformed geometry coordinates of the checkerboard points:
\begin{equation}
	\label{eq:transfer_exp}
	\tilde {\bm P}^{xyz}_c 
	=
	\left [
		\begin{matrix}
			R   & t \\
			0^T & 1 
		\end{matrix}
	\right]
	\bm P^{xyz}_c,
	R \in SO3, t \in R^3
\end{equation}
$\bm I_P$ is the corresponding reflectance intensity value, $\bm S^{xyz}_c$ and $\bm I_S$ are the same representation for the standard model; $N_c$ is the number of points on the checkerboard; $\delta_{in}$ is the discriminant function for judging whether the point $\tilde{p_i}$ falls within the border rectangle of the standard checkerboard, and $G_i$ is the closet corner to $\tilde{p_i}$, as defined in Eq.~\ref{eq:similarity_function_G_d}:

\vspace{-0.2cm}
\begin{equation}
	\begin{aligned}
		\label{eq:similarity_function_G_d}                                        
		G_i = \argmin_{G_j} {|\tilde{p_i} - G_j|},                        
		                                                                          \\
		d( \tilde{p_i}, G_i ) = |G^x_i - \tilde{p^x_i}| + |G^y_i - \tilde{p^y_i}| 
	\end{aligned}
\end{equation}

Thus, the optimal solution of $R, t$ can be solved through nonlinear optimization, the L-BGFS\cite{byrd1995limited} method is utilized for optimizing this problem. Combined with corners $C_{std}$ that are directly generated by the geometric parameter of the checkerboard, the final estimated corner coordinates $C_{3D}$ of the point cloud (Eq.~\ref{eq:corner}) can be obtained through inverse transformation:

\vspace{-0.3cm}
\begin{equation}
	\label{eq:corner}
	\centering
	\begin{matrix}
		C_{3D}
		=
		\left [
		\begin{matrix}
		{R^T} & {-R^T} {t} \\
		{0^T} & 1                  
	\end{matrix}
	\right]
	C_{std}.
	\end{matrix}
	\vspace{-0.4cm}
\end{equation}

\begin{figure*}[!htb]
	\centering
	\subfloat[]{\includegraphics[trim=10 327 200 57, clip, width=0.33\textwidth]{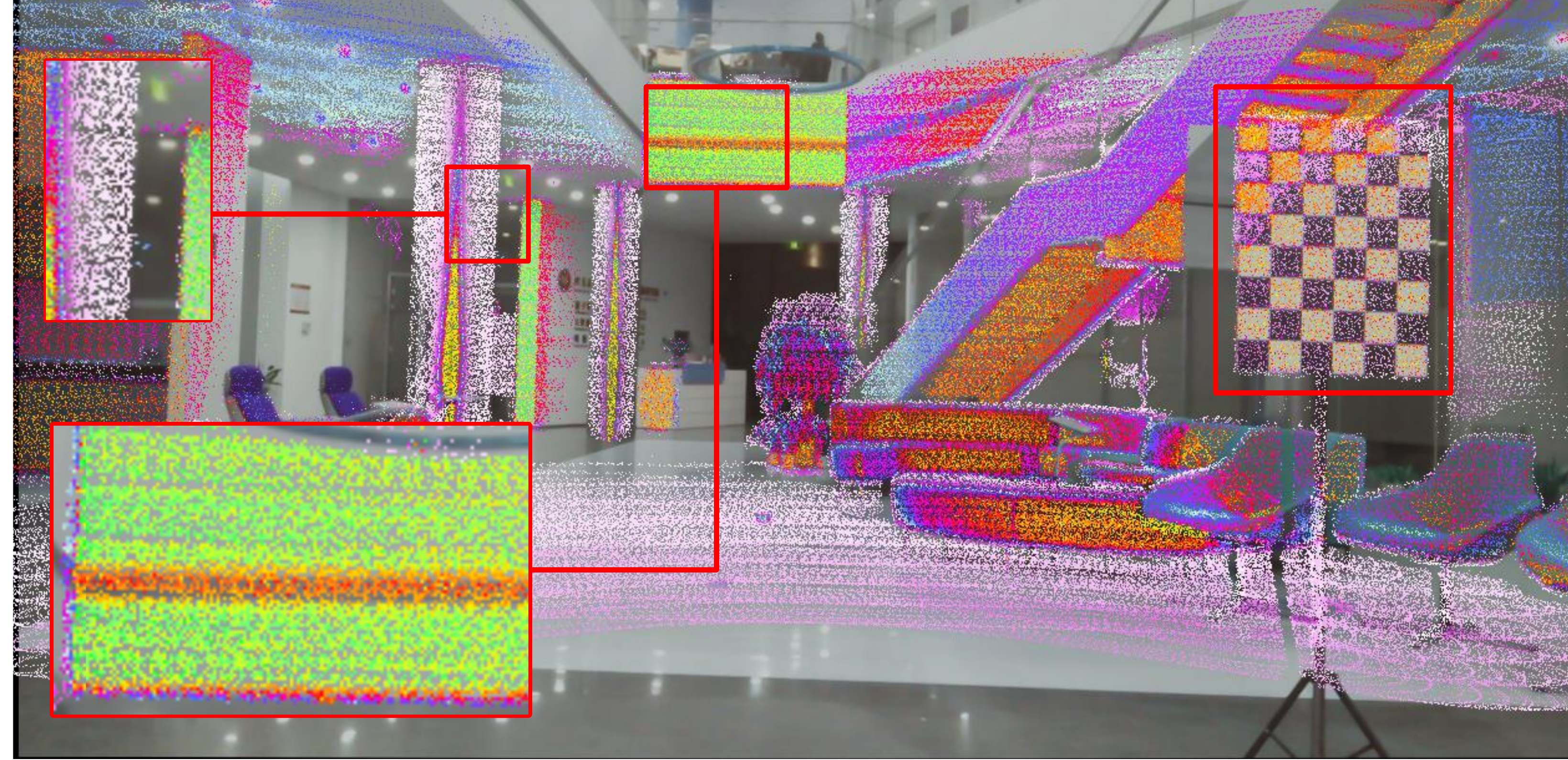}}
	\subfloat[]{\includegraphics[trim=0 105 0 53, clip, width=0.33\textwidth]{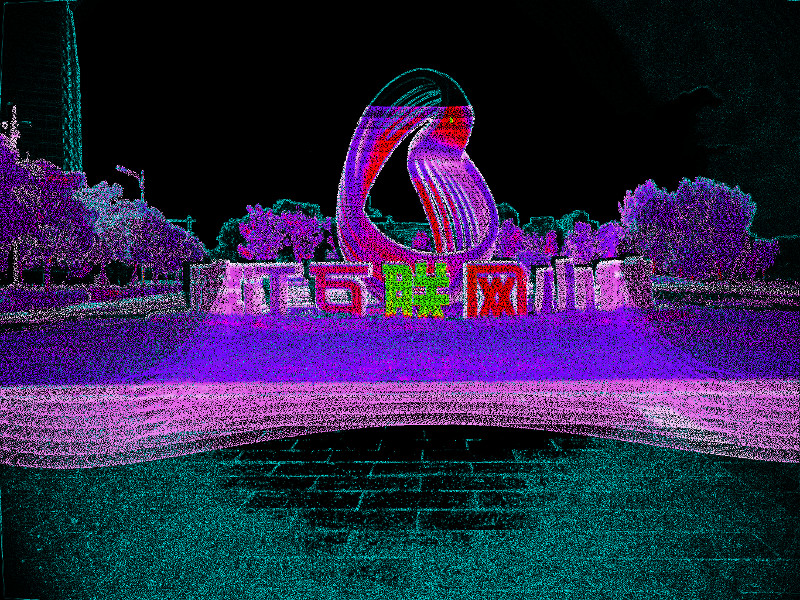}}
	\subfloat[]{\includegraphics[trim=20 160 450 76, clip, width=0.33\textwidth]{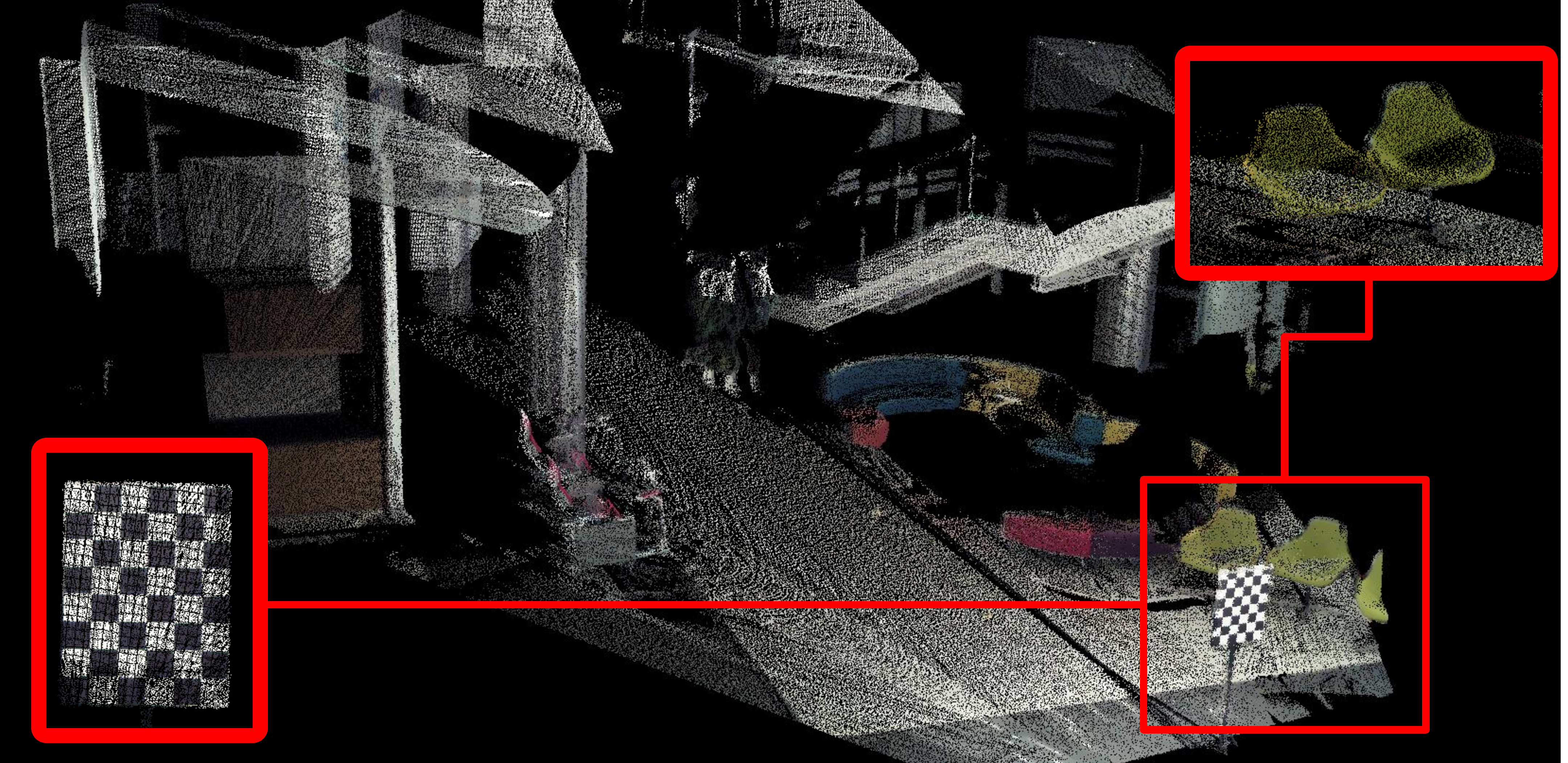}}
	\\[-1.5ex]
	\caption{(a) Visualization of the calibration results. The point clouds are projected onto the image plane based on the solved extrinsic parameter matrix; (b) The projected point clouds on images (with the edge enhanced) in outdoor environment; (c) The colorized point clouds by projecting the pixel from the image using the calibration result.}
	\label{fig:visualization} 
	\vspace{-0.4cm}
\end{figure*}

\begin{figure}[H]
	\centering
	\includegraphics[trim=3 0 0 0, clip, width=0.46\textwidth]{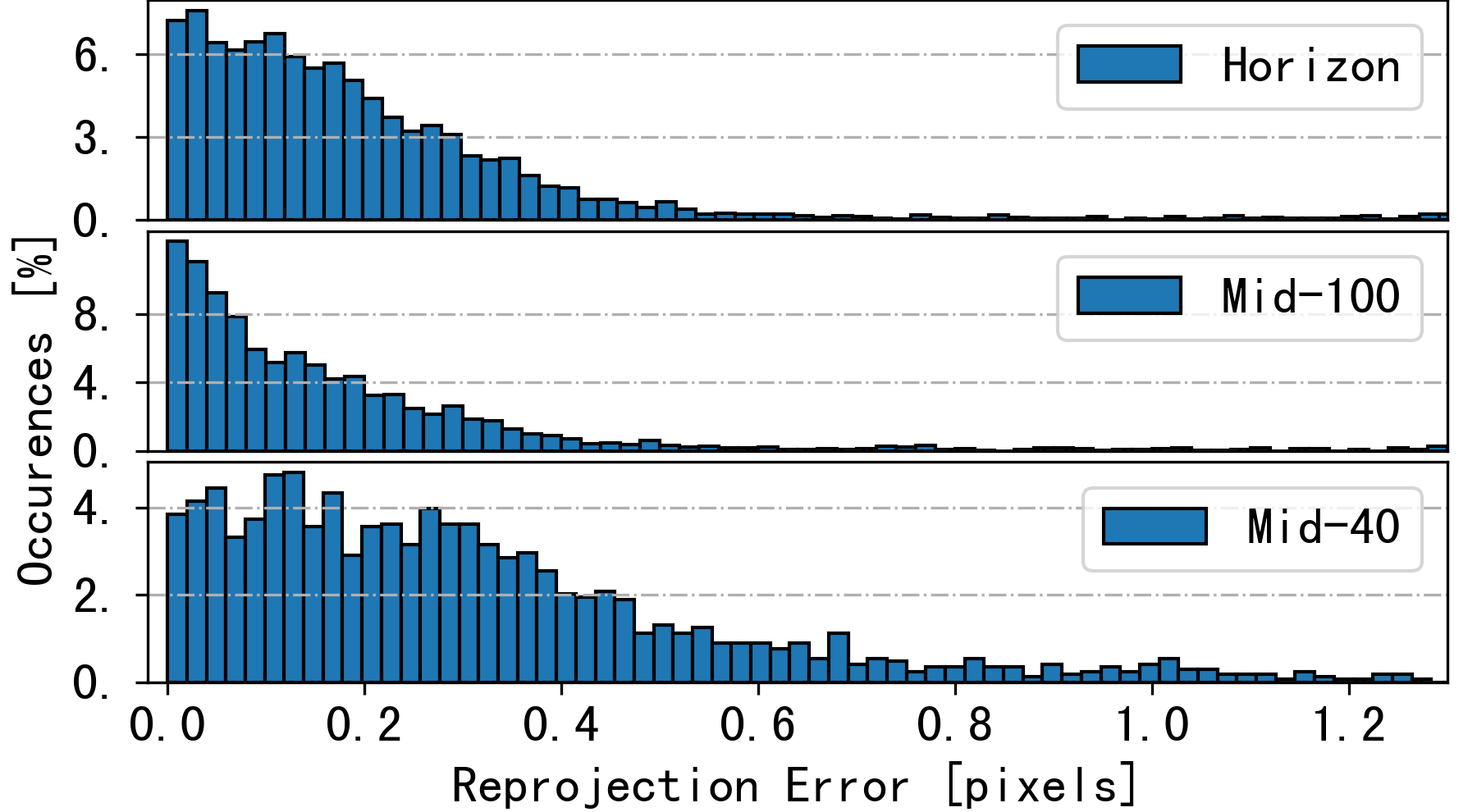} 
	\caption{Normalized reprojection error to evaluate performance of extrinsic calibration of different SSL models. }
	\label{fig:reprojection_error} 
\end{figure}

\subsection{Extrinsic Calibration}

\textbf{2D corner detection from image.} The checkerboard corner detection algorithm \cite{duda2018accurate} is used to detect the corners $C_{2D}$ from the image, which is corresponding to integrated point cloud frame from SSL. Note that due to the symmetry of the checkerboard along the diagonal direction, the order of the corner points detected in the image and in the point cloud may be ambiguous. We reorder the detected corner points and index them from the lower left corner.

\textbf{Iteratively extrinsic parameter solving.} Considering that the limited samples of checkerboard measurement, for the obtained 3D-2D corners, the RANSCA-based PnP is first adopted to get a initial extrinsic solution $\bm {E_0}$; then, the corners with high reprojection error (calculated based on $E_0$) are dropped, and the PnP-solving and reprojection evaluating process continues repeatedly, until all errors are lower than $\delta_{reproj}$. Then final extrinsic matrix is denoted as $\bm E$. 

\section{Experiments and results}

\subsection{Calibration Setup}
In real-condition calibration process, the SSL and camera are required to be in a fixed relative position, and the intrinsic of camera are assumed to be known. Through multiple sampling of the checkerboard placed in different positions and orientations within the FOV of the sensors, the corresponding images and multi-frame point cloud are collected, and are then used to calculate the extrinsic parameters.

We evaluate the proposed method on multiple representative SSL and camera models, the sensor details are described in Table.~\ref{tab:sensor_parameter}.

\begin{table}[!htb]
	\renewcommand\arraystretch{1.1}
	\centering
	\begin{tabular}{cccc}
		\toprule[2pt]
		\multirow{2}{*}{\textbf{Sensor type}}              & \multirow{2}{*}{\textbf{Sensor model}} & \multicolumn{2}{c}{\textbf{FOV($^\degree$)}} \\ \cline{3-4} 
		                                          &                  & \textbf{H}     & \textbf{V}    \\
		\midrule
		\multirow{3}{2cm}{ \centering Solid-state \\ LiDAR} & Livox Horizon     & 81.7  & 25.1 \\
		                                          & Livox Mid-40     & 38.4  & 38.4 \\
		                                          & Livox Mid-100    & 98.4  & 38.4 \\
		\midrule
		\multirow{3}{*}{Camera}
		                                          & MYNT EYE-D D1000 & 103.0 & 55.0 \\
		& PointGrey Flea3          & 111.0 & 82.9  \\
		\bottomrule[2pt]
	\end{tabular}
	\caption{The detailed model and parameter of sensors we used to evaluate the proposed calibration method.}
	\label{tab:sensor_parameter}
	\vspace{-0.1cm}
\end{table}


\subsection{Reprojection Visualization}

To visualize the calibration results, we project the point cloud (collected from both indoor and outdoor environments) onto the image plane with the solved extrinsic parameter matrix, as shown in Fig.~\ref{fig:visualization} (a)-(b), with the colors of the projected points generated according to the corresponding reflectivity. For clarity, the contour is extracted on the image to make the matching difference in the projection results more significant in Fig.~\ref{fig:visualization} (b). It can be seen (especially from the partial enlarged view) that the projected points (with different reflectivity) matches the original image precisely. Fig.~\ref{fig:visualization} (c) shows an example of projecting the 2D image pixels back into the 3D space (also called point cloud colorization\cite{liu2019pccn}), it can be seen that the projected image texture is also accurately matched to the 3D point cloud.

\subsection{Normalized Reprojection Error}

In real conditions, it is hard to get a reference ground-truth measurement of the extrinsic parameters between LiDAR and camera. A compromise solution is to evaluate the result by measuring the errors between the reprojected checkerboard corners from the point cloud and from the image\cite{koide2019general}. We first project the estimated 3D corners of checkerboard $C_{3D}$ to 2D space, based on the camera intrinsic $\bm K$ and estimated extrinsic $\bm E$. Then, the corners from image $C_{2D}$ are treated as pseudo ground-truth to calculate reprojection error. Notice that, due to the perspective projection of camera, the distance of the checkerboard placements may affect the scales of reprojected points (for instance, at the placements that are too far, the reprojection error is smaller than the actual value), therefore, we re-scale the reprojection by distance normalization to reduce this bias, as described in Eq.~\ref{eq:pe1}:

\vspace{-0.1cm}
\begin{equation}
	\label{eq:pe1}
	\centering
	NRE=\sum_{p \in C'_{3D}, c_{2d} \in C_{2D}}{\frac{d(p)}{d_{max}}(|p-c_{2d}|)}
\end{equation}


\begin{equation}
	\label{eq:pe2}
	\centering
	C'_{3D}
	= 
	\frac{1}{z} 
	\bm{E} 
	\bm{K}
	\left [
		\begin{matrix}
			\bm {C^x_{3D}} \\
			\bm {C^y_{3D}} \\
			\bm {C^z_{3D}} 
		\end{matrix}
	\right]
\end{equation}

Here, $c_{2d}$ is the closest image pixel to the reprojected point $p$, and $d(p)$ is the distance of $p$ from LiDAR center. Fig.~\ref{fig:reprojection_error} illustrates the normalized reprojection errors of different LiDAR models, it can be seen that most reprojection errors are within 0.6 pixel, which proves the accuracy of the proposed method. For Fig.~\ref{fig:reprojection_error} (c), the reprojection error of Mid-40 is a slightly higher than that of the other two types, this is because it has a smaller sensor FOV (about 40 degree), thus less target placements can be sampled, causing absent constraints when solving PnP problem. As a result, the calculation accuracy of extrinsic parameters decreases.  

\begin{table}[H]
	\centering
		\begin{tabular}{cccccc}
			\toprule[2pt]
			\multicolumn{1}{c}{\multirow{2}{*}{Method}} & \multicolumn{1}{c}{\multirow{2}{*}{$AVG$}} & \multicolumn{4}{c}{$ NRE<pe~(\%) $}                                                                             \\ \cline{3-6} 
			\multicolumn{1}{c}{}                        & \multicolumn{1}{c}{}                         & \multicolumn{1}{c}{$pe=0.5$} & \multicolumn{1}{c}{$=1$} & $=5$ & $=10$ \\
			\midrule
			MI\cite{pandey2015automatic}                                          & 9.77		&		1.89		&		3.93		&		39.34		&		60.15               \\
			ILCC\cite{wang2017reflectance}                                        &   4.76		&		5.64		&		10.60		&		59.40		&		90.60               \\
			HUANG\cite{huang2020improvements}                                       & 7.73		&		4.54		&		9.41		&		39.33		&		70.59\\
			\textbf{Ours}                                        						& \textbf{ 2.11 }		&		\textbf{ 69.33 }		&		\textbf{ 75.41 }		&		\textbf{ 87.16 }		&		\textbf{ 92.75 }            \\
			\bottomrule[2pt]
		\end{tabular}
	\caption{Quantitative evaluation of the calibration methods. $AVG$ denotes the average normalized reprojection error (pixels); $NRE<pe$ denotes the percentage of corners with reprojection error less than $pe$ pixels.}
	\label{tab:compare}
\end{table}

Table. \ref{tab:compare} shows the quantitative evaluation results of the proposed method and previous studies on the SSL-camera calibration task. Notice that, due to the heavily dependent on the ring-based scanning pattern of mechanical LiDAR (e.g. Velodyne HDL64), most of the implementations of previous studies(\cite{wang2017reflectance, huang2020improvements}) failed to detect the calibration target from SSL point clouds, and therefore refused to calculate extrinsic parameter; only after we resample the integrated points to 128-channel LiDAR-like pattern, can they generate feasible solutions. The proposed method performs more accurate calibration results than the others by a large margin, especially for $pe<0.5$ and $pe<1$, due to the consideration of characteristics of SSL. 

\vspace{-0.1cm}
\begin{figure}[H]
	\centering
	\includegraphics[trim=0 15 15 15, clip,width=0.47\textwidth]{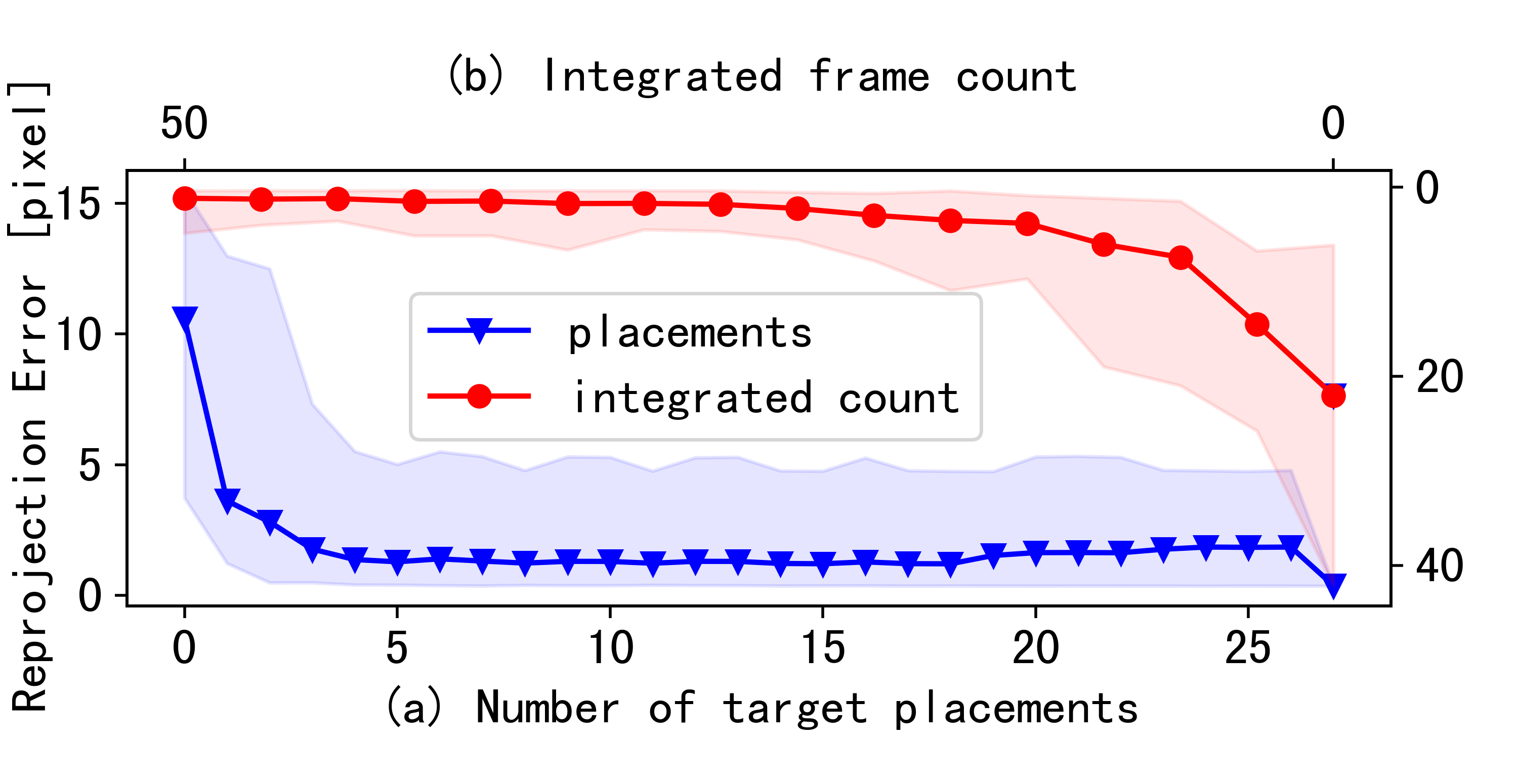} 
	\caption{Impact of different target placements and integrated frame count on the calibration performance. }
	\label{fig:placement} 
	\vspace{-0.1cm}
\end{figure}


We also investigate the influencing factors that affect the reprojection error. As show in Fig.~\ref{fig:placement} (a), the proposed method achieve better performance by adding samples of target placements with different distances and poses, and the error can remain stable with a minimal samples at 5\textasciitilde6 poses. Fig.~\ref{fig:placement} (b) shows that, increasing the integration time can also improve the calibration performance, which verifies the effect of time-domain integration.


\begin{figure}[!htb]
	\centering
	\includegraphics[trim=0 5 0 30, clip, width=0.5\textwidth]{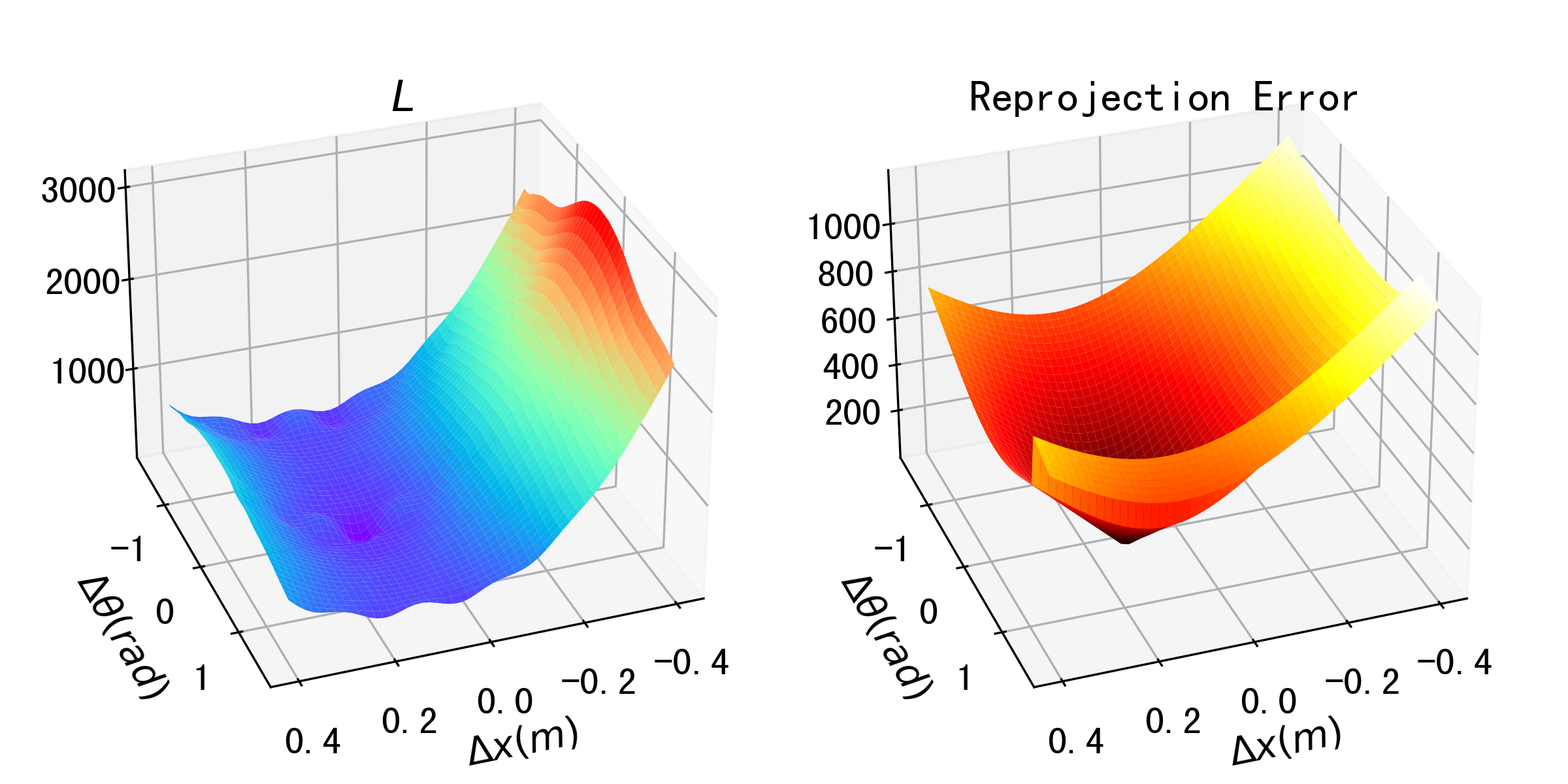} 
	\caption{Distribution of cost function $\mathcal{L}$ and reprojection error with the increments of optimizing variables, i.e. the rigid body transformation parameters between calibration target measurement and standard calibration board model.}
	\label{fig:cost} 
	\vspace{-0.0cm}
\end{figure}

\subsection{Evaluation of 3D Corner Estimation}
\begin{figure}[!htb]
	\centering
	\includegraphics[trim=0 30 0 40, clip, width=0.48\textwidth]{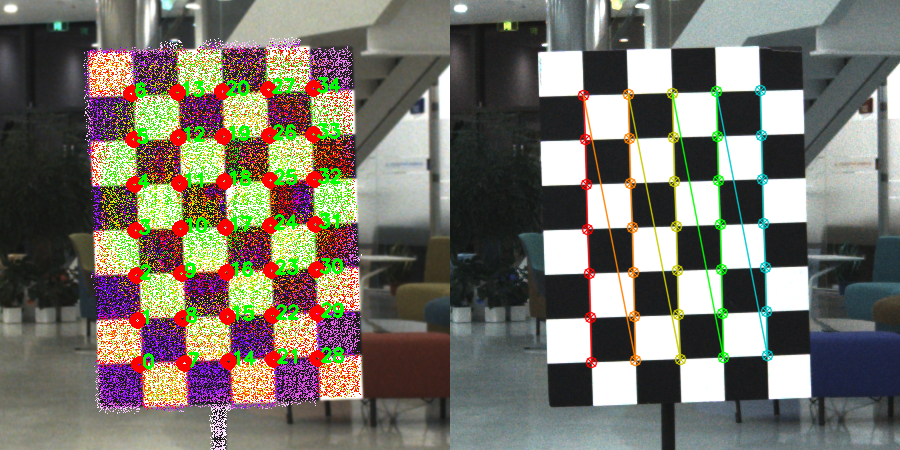}
	\\
	\caption{Visualization the estimated corners of the calibration target. The left is the reprojection of points from SSL, the red dots are reprojection of detected 3D corners; the right is detected 2D corners directly from image. }
	\label{fig:corner_reprojection}  
\end{figure}

The performance of the proposed method is highly related to the estimated checkerboard corners. Therefore, we further discuss the performance of corner estimation. Fig.~\ref{fig:cost} illustrates the distribution of cost function $\mathcal{L}$ and reprojection errors during corner point estimation. The X and Y axes in each subplot donates partial dimensions of the optimization variables (i.e., the transformation between the $\bm{P^I_c}$ and the standard checkerboard model $\bm{S_c}$), the left column is the distribution of similarity function $\mathcal{L}$, and the right column is that of the reprojection error. It can be seen, that in the space which the optimization variables spin, the minimum of $\mathcal{L}$ and the reprojection error are corresponding to the similar region, which means that, as long as a set of optimal solutions for 3D corners are found, the optimal extrinsic parameters can be solved correspondingly. This can prove the rationality of the proposed method from one aspect. Fig.~\ref{fig:corner_reprojection} is the visualization result of the checkerboard corner detected from point cloud and image, respectively, it also shows the consistency of estimated corners both from 2D and 3D space. More detailed experiments can be found on our project website (link in the abstract).


\section{Conclusion}

This paper proposes an novel extrinsic calibration method for SSL-camera systems. With the proposed time-domain integration and feature refinement pipeline, effective information from fuzzy LiDAR measurement can be extracted; based on the reflectance distribution of calibration target point clouds, we proposed a 3D corner estimation from checkerboard measurement, and combine with the 2D corner extracted from the images, the extrinsic parameters are solved by the proposed calibration method. The whole workflow is fully automated, only needs the user to change the position of the checkerboard several times. The extensive experiments demonstrate that our method can perform accurate calibration in real-world conditions. Future work includes conducting a more systematic analysis of the calibration process, and expanding the method into dynamic scenarios.

\bibliography{reference}
\bibliographystyle{unsrt}

\end{document}